%% file: sample-sigconf.tex
\documentclass[sigconf]{acmart}

\usepackage{algorithm, algorithmic}
\usepackage{units}
\usepackage{balance}

\pdfoutput=1
\AtBeginDocument{%
  \providecommand\BibTeX{{%
    \normalfont B\kern-0.5em{\scshape i\kern-0.25em b}\kern-0.8em\TeX}}}

\copyrightyear{2021} 
\acmYear{2021} 
\setcopyright{acmcopyright}\acmConference[CIKM '21]{Proceedings of the 30th ACM International Conference on Information and Knowledge Management}{November 1--5, 2021}{Virtual Event, QLD, Australia}
\acmBooktitle{Proceedings of the 30th ACM International Conference on Information and Knowledge Management (CIKM '21), November 1--5, 2021, Virtual Event, QLD, Australia}
\acmPrice{15.00}
\acmDOI{10.1145/3459637.3481944}
\acmISBN{978-1-4503-8446-9/21/11}

\begin{document}


\title[Profiling Blocks and Design Spaces]{Profiling Neural Blocks and Design Spaces \\ for Mobile Neural Architecture Search}

\author{Keith G. Mills$^{1*\dagger}$, Fred X. Han$^{2*}$, Jialin Zhang$^3$, Seyed Saeed Changiz Rezaei$^2$}
\author{Fabian Chudak$^2$, Wei Lu$^2$, Shuo Lian$^3$, Shangling Jui$^3$, Di Niu$^1$}
\thanks{$^*$Equal contribution}
\thanks{$\dagger$Work done during an internship at Huawei Technologies Canada. \\ Correspondence to: \\Keith G. Mills (kgmills@ualberta.ca), \\Fred X. Han (fred.xuefei.han@huawei.com)}
\affiliation{$^1$University of Alberta \city{Edmonton} \state{AB} \country{Canada}}
\affiliation{$^2$Huawei Technologies \city{Edmonton} \state{AB} \country{Canada}}
\affiliation{$^3$Huawei Kirin Solution \city{Shanghai} \country{China}}

\renewcommand{\shortauthors}{Mills and Han, et al.}

\begin{abstract}
\input{src/abstract}
\end{abstract}

\keywords{Neural Architecture Search; Design Space; Latency Measurement}


\maketitle

\input{src/introduction}
\input{src/background}
\input{src/analysis_b}
\input{src/analysis_ul}
\input{src/application}

\input{src/related}
\input{src/conclusion}

\bibliographystyle{ACM-Reference-Format}
\balance
\bibliography{sample-base}


\end{document}

%% file: src/abstract.tex
Neural architecture search automates neural network design and has achieved state-of-the-art results in many deep learning applications. While recent literature has focused on designing networks to maximize accuracy, little work has been conducted to understand the compatibility of architecture design spaces to varying hardware. In this paper, we analyze the neural blocks used to build Once-for-All (MobileNetV3), ProxylessNAS and ResNet families, in order to understand their predictive power and inference latency on various devices, including Huawei Kirin 9000 NPU, RTX 2080 Ti, AMD Threadripper 2990WX, and Samsung Note10. We introduce a methodology to quantify the friendliness of neural blocks to hardware and the impact of their placement in a macro network on overall network performance via only end-to-end measurements. Based on extensive profiling results, we derive design insights and apply them to hardware-specific search space reduction. We show that searching in the reduced search space 
generates better accuracy-latency Pareto frontiers than searching in the original search spaces, customizing architecture search according to the hardware. Moreover, insights derived from measurements 
lead to notably higher ImageNet top-1 scores on all search spaces investigated.

%% file: src/introduction.tex
\section{Introduction}
\label{sec:intro}

Neural Architecture Search (NAS) has been established as the de facto method for automating neural network design~\cite{liu2018DARTS, cai2018proxylessnas,  chen2019progressive}, leading to state-of-the-art models in computer vision, especially backbone networks that achieve top performance on the ImageNet~\cite{russakovsky2015imagenet} benchmarking dataset~\cite{tan2019efficientnet, cai2020once, bender2020can}.
Hardware-aware NAS has recently attracted much attention in both academia and industry, aiming at finding high-performing and hardware-friendly architectures tailored to different platforms,
ranging from fleets of industry-grade GPUs~\cite{zoph2018learning, yu2020bignas} and consumer-grade graphics cards~\cite{xu2020pcdarts} to mobile processors~\cite{zhang2020fast} and IoT devices~\cite{lin2020mcunet}.  

A majority of NAS research has focused on innovating the search strategies, including those based on Evolutionary Computing~\cite{real2019regularized}, Reinforcement Learning~\cite{zoph2017NAS}, Gradient Descent~\cite{liu2018DARTS}, and Bayesian Optimization~\cite{white2019bananas}, etc., as well as newer and more complex search algorithms~\cite{chen2020stabilizing, li2020geometry, peng2020cream}. However, a common observation is that state-of-the-art architectures produced by NAS \cite{tan2019efficientnet, cai2020once, bender2020can} still critically depend on the choice of the architecture search space (or \textit{design space}~\cite{radosavovic2019network, radosavovic2020designing, you2020design}). There are several top-performing and frequently used families of design spaces, including MobileNetV2~\cite{sandler2018mobilenetv2}, MobileNetV3~\cite{howard2019searching}, ResNet~\cite{he2016deep}, etc., each of which is built upon a set of predefined neural blocks, e.g., MBConv blocks \cite{howard2017mobilenets} for MobileNets. 

\begin{figure*}[t]
	\centering
	\includegraphics[width=7in]{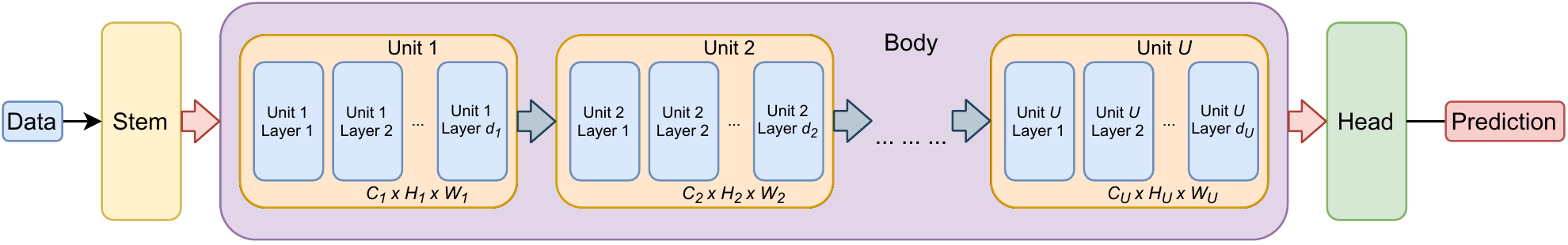}
	\vspace{-5mm}
	\caption{The high-level macro layout of the neural networks analyzed, which represents many deep architectures. The body of each network consists of $U$ units, each containing $d_u$ layers and operating on a unique tensor shape $C_u\times H_u\times W_u$. A neural block (i.e., an operation) is placed at each layer.}
	\vspace{-3mm}
	\label{fig:netLayout}
\end{figure*}

Although a number of state-of-the-art networks, e.g., EfficientNet~\cite{tan2019efficientnet}, Once-for-All~\cite{cai2020once}, TuNAS~\cite{bender2020can}, etc., have been discovered by combining and arranging these popular neural blocks, little has been done to understand the predictive power of these neural blocks as building blocks of a high performing network, or their friendliness to different hardware devices, especially in terms of inference latency. However, such an understanding is desirable as there is an imperative need to deploy customized neural networks onto each type of hardware instead of merely increasing accuracies on GPUs yet at the cost of large models. A prior knowledge of neural block performance and their compatibility with hardware will enable a search algorithm to focus on important regions of a design space, reducing search cost and improving end results.

In this paper, we propose a neural network profiling method for gauging the impact that popular neural blocks and their placement in a macro structure have on the overall network performance in terms of prediction accuracy, latency, FLOPS, etc. on a number of hardware devices, with a focus on their behavior on mobile hardware as compared to GPU/CPU. We aim to obtain insights from neural block profiling such that hardware dependent search space reduction and pruning can be performed to enhance search. 

Specifically, we propose a method for profiling the neural blocks used in the state-of-the-art mobile architecture design spaces of Once-for-All (OFA)~\cite{cai2020once} and ProxylessNAS~\cite{cai2018proxylessnas} as well as a variant of the well-known ResNet50~\cite{he2016deep}. Although typically only the performance metric values pertaining to an entire network can be measured, we introduce a sampling-based method that is able to quantify the effect of specific blocks as well as their placement position in a macro net.
In order to quantify the contribution from each block to the whole network, our method involves randomly sampling architectures from a given design space, while fixing a specified neural block (aka operation) to a certain layer in the macro net. We then measure the end-to-end performance of each network to produce a sample distribution that can quantify the aggregate effect of a specific block placement.

A highlight of our contribution is that we perform extensive latency profiling of neural blocks in the selected design spaces across multiple devices including mobile devices such as Huawei Kirin 9000 NPU, Samsung Note10,
as well as Nvidia RTX 2080 Ti GPU and AMD Threadripper 2990WX CPU. Based on both accuracy profiling and hardware-specific metrics, We discover and offer insights into the behaviour of the neural blocks on these different hardware platforms. We quantify effects that different blocks have on overall performance metrics when present in a network. We also profile how sensitive network topology is to block choice and network depth.
These knowledge and insights will help to rule out unpromising regions of a search space during hardware-aware NAS and boost search effectiveness.

We demonstrate the utility of the insights and knowledge discovered from profiling by applying them for \textit{a priori} design space reduction and pruning before search. We execute searches on both the original and reduced search spaces and show that the resultant Pareto frontiers found in our insight-driven search spaces outperform those found in the original spaces. Additionally, the derived insights when used with a simple evolutionary algorithm allow us to notably outperform OFA$_{Large}$, the best architecture originally found on OFA-MobileNetV3 by NAS in terms of ImageNet top-1 accuracy. 

Finally, the networks on the Huawei-NPU Pareto frontier found by insight-driven search on OFA have been used as backbone networks in several Kirin products, including a Hi-AI tracking task to improve the motion auto-focusing of the camera, especially on fast-moving objects. The adoption of these networks leads to latency and model size reduction and have been deployed in production since March 2021. The data collected to perform this study will be released\footnote{Code and model data published at https://github.com/Ascend-Research/BlockProfile} together with the analysis code to facilitate future research on hardware-friendly NAS.

%% file: src/background.tex
\section{Background}
\label{sec:background}

In this section, we describe the families of search spaces considered in our measurement study, including OFA-MobileNetV3 (or OFA in short), ProxylessNAS and ResNet50, in terms of the macro network, topologies, blocks, operations used and input resolutions adopted. We also provide technical details how data collection is performed.

\subsection{Design Spaces}
\label{sec:design_spaces}

A high-level view of the network topologies considered in this paper is given in Figure~\ref{fig:netLayout}. Networks consist of stacking multiple \textit{units}, each corresponding to a specified latent tensor dimension, together to form the searchable body. Each unit consists of a variable number of \textit{layers}, and each layer contains a single block chosen from a set of candidate blocks. The convolution operation corresponding to the block residing in the first layer of a given unit has a stride of $2$ in order to halve the height and width of latent data while increasing the number of channels. Table~\ref{table:operations} enumerates all candidate blocks. We consider three design spaces, described as follows.

\begin{table*}[t]
    \vspace{-5mm}
	\centering
	\caption{Candidate blocks for MobileNets (OFA and ProxylessNAS; left) and ResNet50 (right).
	Blk. Code is a proxy name we use for figures in Section~\ref{sec:full_span} to simplify notations.
	}
	\label{table:operations}
	\begin{tabular}{lccc|lccc} 
	\toprule
	    \textbf{MobileNets} & \textbf{Exp. Ratio} & \textbf{Kernel Size} & \textbf{Blk. Code} & \textbf{ResNet50} & \textbf{Unit Ratio} & \textbf{Layer Ratio} & \textbf{Blk. Code} \\ \midrule
	    MBConv3--3 & 3 & 3$\times$3 & B1 & 65--0.20 & 0.65 & 0.20 & C65--B20\\
	    MBConv3--5 & 3 & 5$\times$5 & B2 & 65--0.25 & 0.65 & 0.25 & C65--B25\\
	    MBConv3--7 & 3 & 7$\times$7 & B3 & 65--0.35 & 0.65 & 0.35 & C65--B35 \\
	    MBConv4--3 & 4 & 3$\times$3 & B4 & 80--0.20 & 0.8 & 0.20 & C80--B20 \\
	    MBConv4--5 & 4 & 5$\times$5 & B5 & 80--0.25 & 0.8 & 0.25 & C80--B25 \\
	    MBConv4--7 & 4 & 7$\times$7 & B6 & 80--0.35 & 0.8 & 0.35 & C80--B35 \\
	    MBConv6--3 & 6 & 3$\times$3 & B7 & 100--0.20 & 1.0 & 0.20 & C100--B20 \\
	    MBConv6--5 & 6 & 5$\times$5 & B8 & 100--0.25 & 1.0 & 0.25 & C100--B25 \\
	    MBConv6--7 & 6 & 7$\times$7 & B9 & 00--0.35 & 1.0 & 0.35 & C100--B35 \\
        \bottomrule
	\end{tabular}
\end{table*}

\textbf{OFA-MobileNetV3 (OFA).} The search space of Once-for-All \cite{cai2020once}, as originally introduced, consists of \textit{MBConv} blocks in MobileNetV3~\cite{howard2019searching}, which have been used to construct multiple state-of-the-art architectures, including TuNAS~\cite{bender2020can} and BigNAS~\cite{yu2020bignas}.
Each \textit{MBConv} block consists of a linear bottleneck, after which the channels are multiplied by an expansion ratio. A depth-wise convolution with a specified kernel size is performed, then followed by a Squeeze-and-Excite~\cite{hu2018squeeze} operation before another bottleneck reduces the number of channels back to the original number. The input is then added to the output. Finally, blocks in the later half of the network use the \textit{h-swish} activation function instead of the traditional ReLU. OFA consists of 5 units, each containing 2--4 layers, depending on the selected architecture. Additionally, OFA incorporates the ability to accept input images of varying resolutions other than the standard 224 pixels squared. For the purposes of this paper we will focus on three resolutions for OFA, $\{192, 208, 224\}$.

\textbf{ProxylessNAS} search space consists of \textit{MBConv} blocks from MobileNetV2~\cite{sandler2018mobilenetv2}, which relies exclusively on the ReLU nonlinearity and lacks the Squeeze-and-Excite operation used in V3. In terms of topology, ProxylessNAS is similar to OFA, with the exception of a sixth unit that always contains a single layer. 

\textbf{ResNet50} is a classical architecture from before MobileNets emerged. Unlike OFA and ProxylessNAS, the number of channels in each unit of ResNet50 is variable and determined by multiplying the maximum number of channels per unit by one of three ratios, $\{0.65, 0.8, 1.0\}$. The blocks in each layer have expansion ratios chosen from $\{0.2, 0.25, 0.35\}$. All blocks are residual; the input is added to the output prior to the final activation. Additionally, all layers consist of three sets of convolution operations. The first and final convolutions have a kernel size of 1, while the second has a size of 3. The overall search space consists of 4 units, the first, second, and final ones containing 2--4 layers, while the third contains 4--6 layers.

\begin{figure*}[t]
	\centering
	\includegraphics[width=7in]{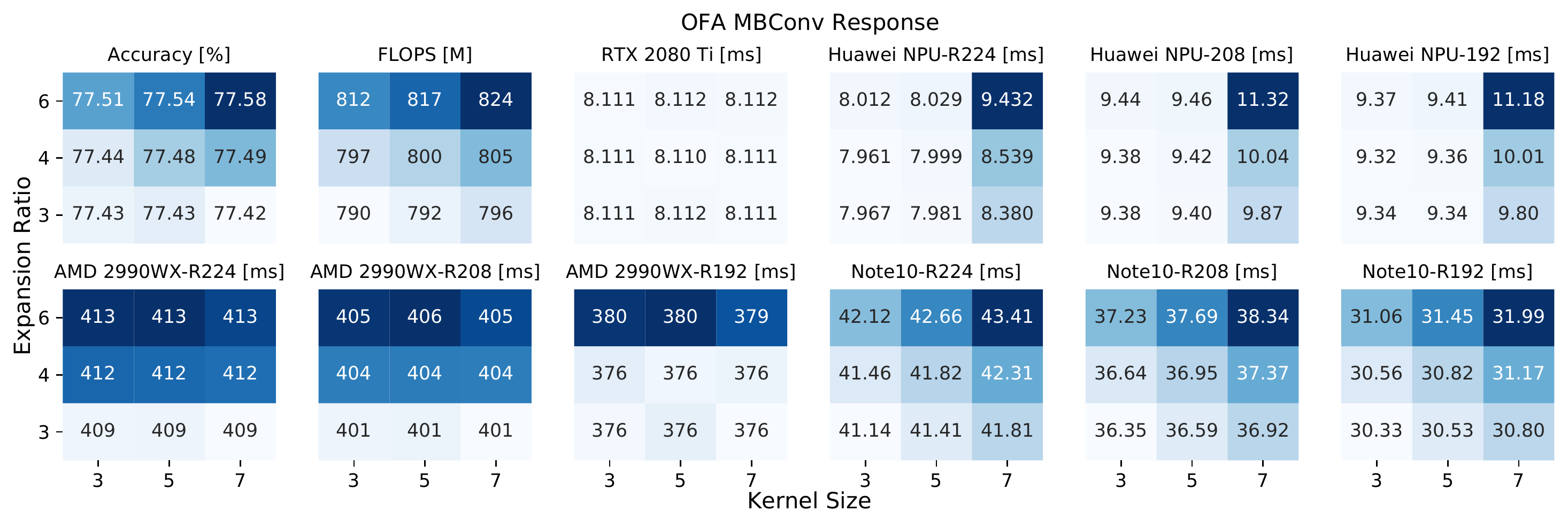}
	\vspace{-5mm}
	\caption{Block-wise average response $M_b$ for OFA-MobileNetV3 blocks in terms of accuracy, FLOPS, and latency on 4 hardware devices. Each entry corresponds to a \textit{MBConv} block identified by an expansion ratio and a kernel size. `-R' flags indicate use of a specific input resolution, assuming 224 by default.}
	\label{fig:OFA_OPS}
	\vspace{-3mm}
\end{figure*}

\subsection{Data Collection}
\label{sec:data}

In this work, we mainly consider accuracy, inference latency and FLOPS as performance metrics.
To obtain the accuracy of an architecture sampled from the above three design spaces, we use the OFA repository\footnote{https://github.com/mit-han-lab/once-for-all}, which provides pre-trained supernet model weights\footnote{We use {\tt ofa\_mbv3\_d234\_e346\_k357\_w1.2} for OFA-MobileNetV3}. OFA~\cite{cai2020once} essentially provides a supernet training mechanism to superpose all architectures in each of the above design spaces such that the architecture accuracy based on the supernet weights is close to the true accuracy obtained by training it from scratch, with proved performance \cite{cai2020once}. Accuracy measurement requires evaluation based on supernet weights on the ImageNet~\cite{russakovsky2015imagenet} validation set. In addition, the repository also provides an accuracy predictor for OFA-MobileNetV3. Obtaining measurement from predictors is less costly than from supernets.

We obtain latencies on the Qualcomm Snapdragon 855 processor in the Samsung Note10 by utilizing the predictor provided by the OFA repository. We also train our own GRU-based latency predictors for Huawei Kirin 9000 NPU, Nvidia RTX 2080Ti GPU and AMD Threadripper 2990WX CPU. To collect latency data for the Kirin 9000 NPU, we measure the end-to-end latency of 5k networks on a universal UDP development board with the NPU and record the inference time needed to feed-forward a single image. We perform the same procedure to measure 50k networks on the 2080 Ti GPU and 15k networks on the 2990WX CPU. The collected data is used to train our predictors. We measured FLOPS for ResNet50 using the {\tt torchprofile} package\footnote{https://github.com/zhijian-liu/torchprofile/}.

%% file: src/analysis_b.tex
\section{Block-wise Performance}
\label{sec:blocks}

In this section, we first describe our methodology for quantifying the general predictive power or resource cost of a given neural block. We then present the average performance of each block pertaining to different hardware devices for all three design spaces. 

\subsection{Methodology}
\label{sec:blocksTheory}
Measuring the accuracy of a single block is difficult since the block must be used with other blocks in a macro network to achieve a certain accuracy. Similarly, measuring the latency of an individual block on a device is challenging because the latency is affected by the input resolutions, channel sizes and neighboring blocks due to hardware-enabled operation fusion. Despite these factors, we observe that each single block still demonstrates different general predictive power, resource cost and latency, by considering the average behavior of randomly sampled networks using the block. Moreover, their average performance when placed in networks may also be used to assess different hardware behavior.

Formally, let $A$ denote an architecture randomly sampled from one of the search spaces mentioned above, allowing a variable number of layers in each unit. Furthermore, let $A_{u,l,b}$ denote a randomly sampled architecture such that block $b$ is fixed to layer $l$ in unit $u$ while unit $u$ has at least $l$ layers. Both the number of layers per unit and block per layer are uniformly sampled under this constraint. Then, we can measure the end-to-end performance of a network $A_{u,l,b}$ as $M(A_{u,l,b})$, where $M$ is a desired metric.

Therefore, to quantify the average performance $M_b$ of a block $b$ on a given metric $M$, we can obtain the expectation of this metric averaged across all locations $(u,l)$ in the network: 
\begin{equation}
\centering
\label{eq:computeBlock}
M_b = \frac{1}{\sum_{u=1}^{U}d_u}\sum_{u=1}^{U}\sum_{l=1}^{d_u}\mathbb{E}[M(A_{u,l,b})]. 
\end{equation}

It is critical that the sampling of each location $(u, l)$ is uniform for the value of $M_b$ to be reflective of $b$ rather than $(u, l, b)$. 

\subsection{Results and Analysis}
\label{sec:blockAnalysis}

Figure~\ref{fig:OFA_OPS} displays block-wise heatmaps on OFA in terms of supernet accuracy, FLOPS, and latency on different devices, as the expansion ratio and kernel size of \textit{MBConv} vary.
Visually, we note a high correlation between accuracy and FLOPS, as both metrics follow nearly identical increasing trends with block size, but emphasizes expansion ratio slightly more than kernel size.

By contrast, latency on the Huawei Kirin 9000 NPU largely depends on kernel size. Size 7 is distinctly unfriendly as the version of the NPU under consideration has special optimization techniques built-in for computing convolutions with kernel sizes of 3 or 5, but not for size 7. Additionally, contrary to intuition, the resolution corresponding to the lowest latencies is 224, even though it necessitates more computations than 208 or 192. The reason is that 224 is a common resolution for computer vision tasks~\cite{liu2018DARTS, tan2019efficientnet, bender2020can}, one that the Da Vinci 2.0 Architecture of the NPU is designed to accept as a template. When uncommon resolutions like 208 are passed to the NPU, it first pads the data until it matches the size of the next-largest template, which explains the additional latency incurred when using smaller images.

\begin{figure}[t]
	\centering
	\includegraphics[width=2.5in]{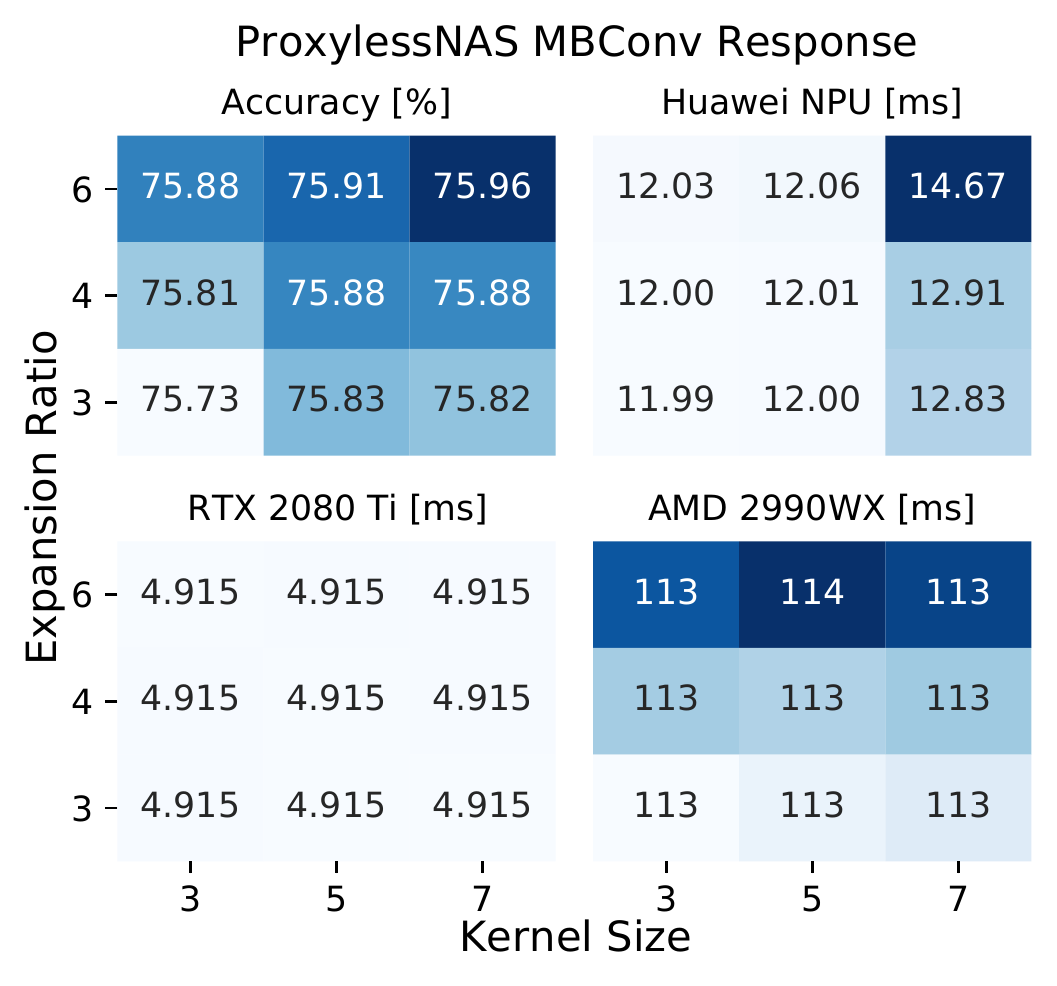}
	\vspace{-3mm}
	\caption{Block-wise average response $M_b$ for blocks in ProxylessNAS on 3 different hardware devices.}
	\label{fig:PN_OPS}
	\vspace{-5.5mm}
\end{figure}

When the NPU can take advantage of multiple optimization techniques, it outperforms Nvidia 2080 Ti. The GPU is the most balanced device in terms of latency as all blocks are equally hardware-friendly. By contrast, the AMD 2990WX CPU is the least hardware-friendly device, with inference latencies ranging in hundreds of milliseconds. CPU latency follows the opposite trend as NPU latency. Instead of increasing dramatically if a specific kernel size is selected, latency largely depends on the expansion ratio. Moreover, CPU latency is dependant on the input resolution. When images of size 224 or 208 are used, the latency of all blocks with expansion ratios of 4 or higher is high and fixed. However, this does not hold when the resolution is further reduced to 192, as the latency of blocks with an expansion ratio of 4 drops to the same values as expansion ratio 3 blocks.

Finally, Samsung Note10 latency rises and falls as a function of both expansion ratio and kernel size, much like accuracy and FLOPS. Additionally, latency decreases linearly as the resolution is shrunk. In contrast to the other mobile processor used, the Huawei NPU, these trends indicate a lack of specific optimizations in favor of a simpler and intuitive performance profile.

Next, Figure~\ref{fig:PN_OPS} displays block profiling results for ProxylessNAS. Compared to OFA, the average accuracy per block is lower, but the accuracy trend is similar with only subtle differences in how close the accuracy of larger blocks, such as MBConv4--7, are to the largest block, MBConv6--7. NPU, GPU and CPU latencies all follow the same trend as OFA, although with different scales. While GPU and CPU latencies are much lower for ProxylessNAS, NPU latency is higher. This is due to differences in the internal structures of MobileNetV3 and ProxylessNAS backbones.
The MobileNetV3 backbone contains more element-wise matrix operations per layer, which is handled well by the NPU but less so by the GPU and CPU.
Meanwhile, the ProxylessNAS backbone has more channels per layer, and their impact on latency is more apparent on an NPU, which is designed for mobile phones with limited memory. 

\begin{figure}[t]
	\centering
	\includegraphics[width=2.5in]{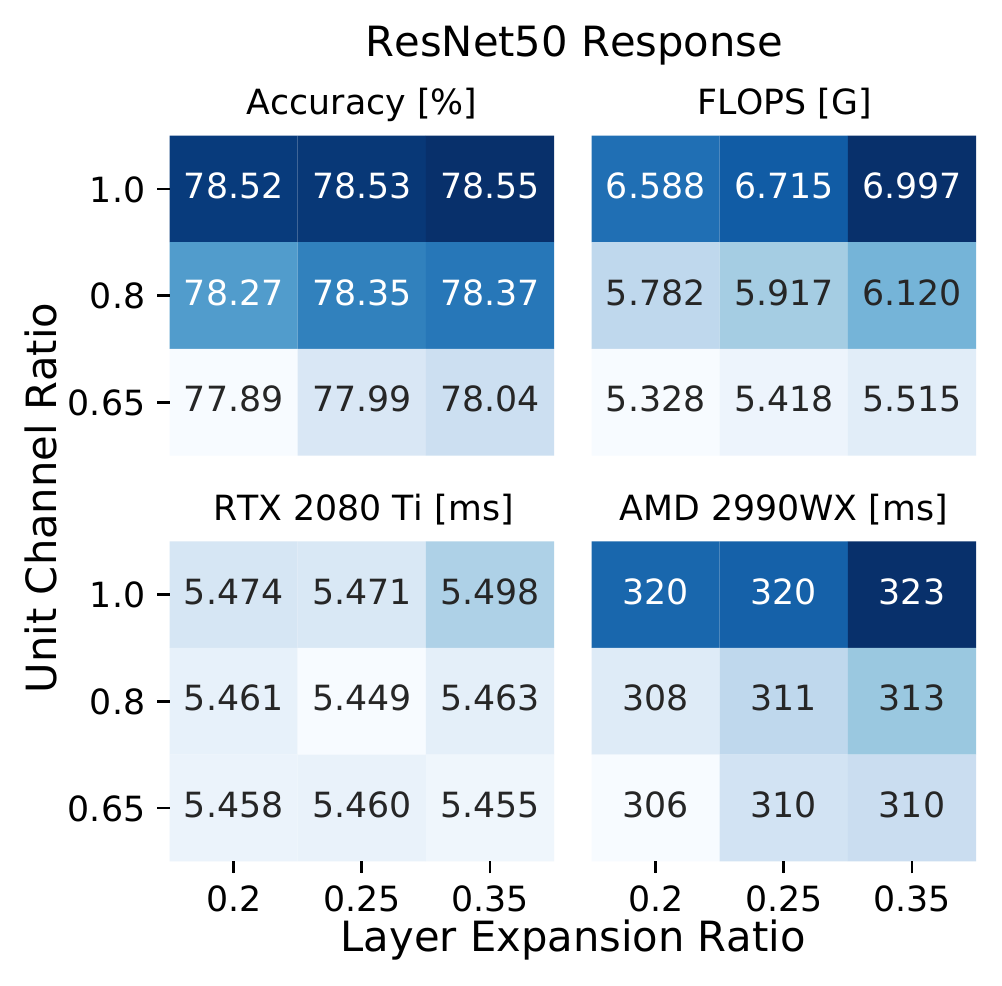}
	\vspace{-3mm}
	\caption{Block-wise average response $M_b$ for blocks in ResNet50 on the GPU and CPU. }
	\vspace{-5mm}
	\label{fig:RN_OPS}
\end{figure}

Furthermore, Figure~\ref{fig:RN_OPS} corresponds to block profiling in ResNet50 for accuracy, FLOPS and latency only on the GPU and CPU, as ResNet is not designed for mobile devices. Block accuracy is the highest for ResNet50 and follows a trend that places significant emphasis on the unit channel ratio over layer expansion ratio. Model size is also much larger and measured in \textit{gigaFLOPS} opposed to \textit{megaFLOPS}. FLOPS on ResNet50 follow a similar trend to FLOPS on OFA, placing roughly equal emphasis on unit and layer ratios.

GPU latency is relatively even, but features a noticeably higher degree of variance than the other search spaces. Finally, the trend of ResNet50 CPU latency resembles a mix of the accuracy and FLOPS trends, with a higher focus on the unit channel ratio.

%% file: src/analysis_ul.tex
\section{Layer-Dependent Performance}
\label{sec:unitlayers}

Since the same block can have a different effect when placed at a different depth in a network, it is necessary to also profile block performance depending on locations of placement $(u, l)$. Doing this will not only assess the relative desirability of blocks, but also allows us to understand which layers/units (whether the early or deeper units) are more sensitive to block variation in terms of accuracy and latency as well as other metrics.
As a result, carefully selecting blocks in these layers could become critical at determining the final performance for a specific hardware. 

Recall from Figure~\ref{fig:netLayout} that each unit within a network uses a different tensor shape. Given this knowledge, we can infer that the interaction of each block with these dimensions will vary. In this section we first describe how to quantify these differences, then we analyze the interaction of block $b$ and placement $(u, l)$ on different hardware.

\subsection{Methodology}
\label{sec:profiling}

Given a design space, recall that $A$ denotes a randomly sampled architecture and $A_{u,l,b}$ denotes a randomly sampled architecture with block $b$ fixed to location $(u,l)$.
To quantify the effect of placing block $b$ at location $(u,l)$ on a certain metric $M$, we consider the \textit{relative expectation} $M_{u,l,b}$, defined as

\begin{equation}
    \centering
    \label{eq:expectation}
    M_{u,l,b} = \mathbb{E}(M(A_{u,l,b})) - \mathbb{E}(M(A)).
\end{equation}
Similarly, we can also compute the \textit{relative $\tau$-percentile} $M_{\tau,(u,l,b)}$ as 
\begin{equation}
    \centering
    \label{eq:quantile}
    M_{\tau,(u,l,b)} = \mathbb{Q}_\tau(M(A_{u,l,b})) - \mathbb{Q}_\tau(M(A)),
\end{equation}
where $\tau \in [0\%, 100\%]$ is the desired percentile, and $\mathbb{Q_\tau}(M(A))$ gives the $\tau$-percentile of a collection of random architectures $A$ in terms of the metric $M$.
This is useful for deciding whether a specific placement $(u,l,b)$ can change the landscape of top-performing or lowest-performing architectures. 

It is worth noting that our method of profiling block placement performance costs far less than profiling an entire search space by exhaustively evaluating all possible architectures as has been done in multiple NAS Benchmarks ~\cite{ying2019nasbench101, dong2020nasbench201}. For example, the OFA design space contains $10^{19}$ distinct architectures~\cite{cai2020once}. 
In contrast, the number of unit-block-layer combinations is 5 $units$ $\times$ 4 $\nicefrac{layers}{unit}$ $\times$ 9 $\nicefrac{blocks}{layer}$, which is 180, which is less than the number of architectures by many magnitudes, and even given multiple samples to compute the expectations, would require far less resources to profile than measuring an entire search space.

\subsection{Results and Analysis}
\label{sec:full_span}

We now profile block sensitivity by quantifying the statistics of $M(A_{u,l,b})$ relative to $M(A)$. Then, we analyze how the number of layers in a unit can influence accuracy and latency.

\begin{figure*}[t]
	\centering
	\includegraphics[width=7in]{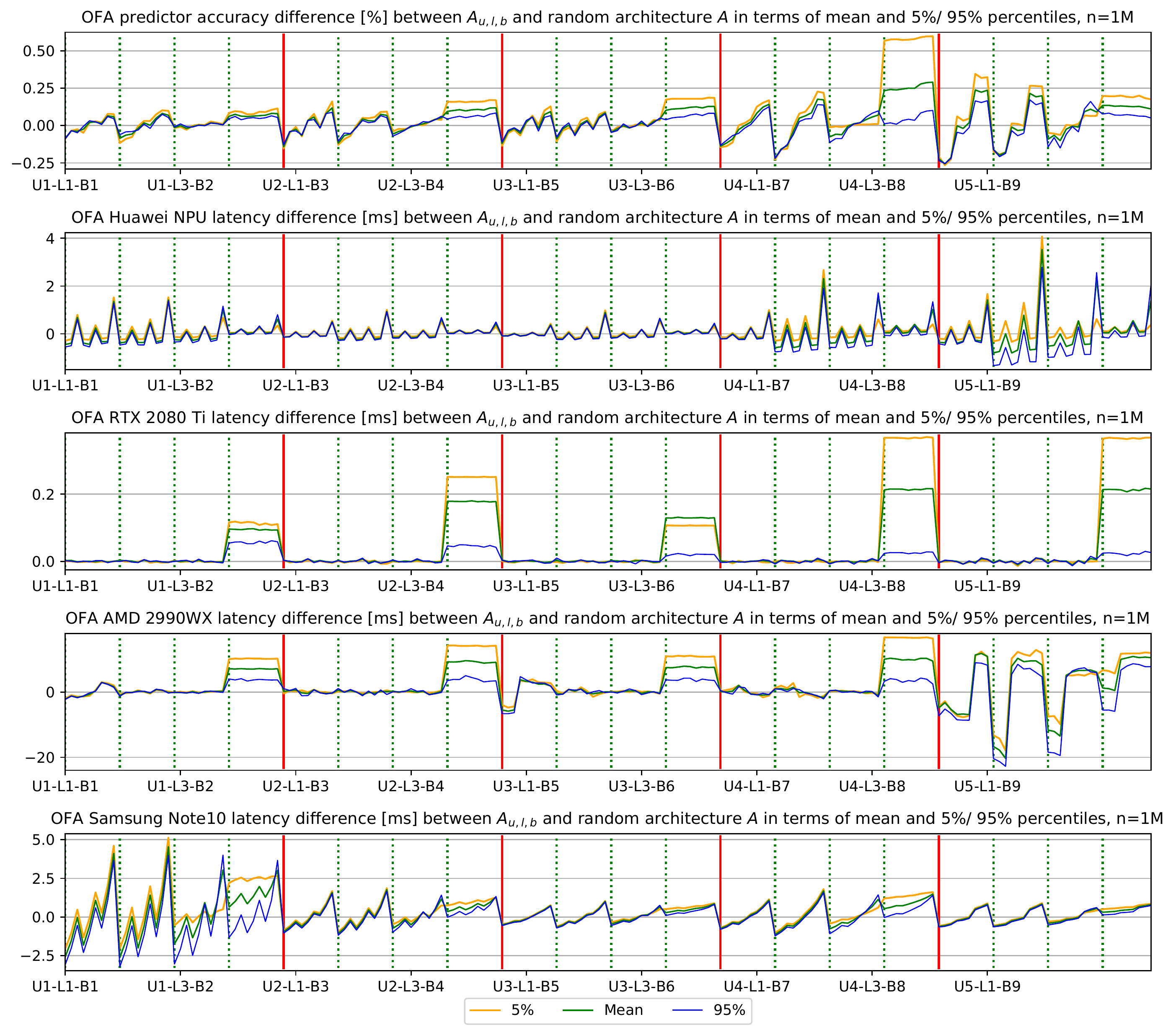}
	\vspace{-8mm}
	\caption{$M_{u,l,b}$ and $M_{\tau, (u,l,b)}$ for OFA-MobileNetV3, where  Unit-Layer-Blk (U-L-B) triplets are arranged in increasing order on x-axis, with solid vertical lines demarcating units and dashed vertical lines demarcating layers.  
	For each metric, we present 3 curves, the mean as well as 5\% and 95\% percentile relative performance according to~\eqref{eq:expectation} and \eqref{eq:quantile}. Input resolution is 224.}
	\vspace{-3mm}
	\label{fig:ofa_sweep}
\end{figure*}

\textbf{Sensitivity to Block selection.} Figure~\ref{fig:ofa_sweep} maps the relative performance for all $(u,l,b)$ combinations for OFA in terms of mean as well as 5\% and 95\% percentile differences on different metrics. For a given layer, we enumerate blocks following the order presented in Table~\ref{table:operations}.

First, we observe that block choice influences accuracy more in the later units, judging by the accuracy difference between the first and last blocks in a given layer. The story is similar for NPU latency. Additionally, latency is more sensitive to block choice in unit 1 than units 2 and 3 because unit 1 processes the tensors with the largest height and width, however across all units it is clear that kernel size 7 is distinctly unfriendly and is an obvious choice for removal when crafting a reduced search space.

By contrast, GPU latency is not sensitive to block choice while CPU latency in unit 5 oscillates according to expansion ratio. Lastly for OFA, Samsung Note10 latency is most responsive in unit 1, while steadily losing sensitivity in units 2 and 3.

\begin{figure*}[t]
	\centering
\includegraphics[width=7in]{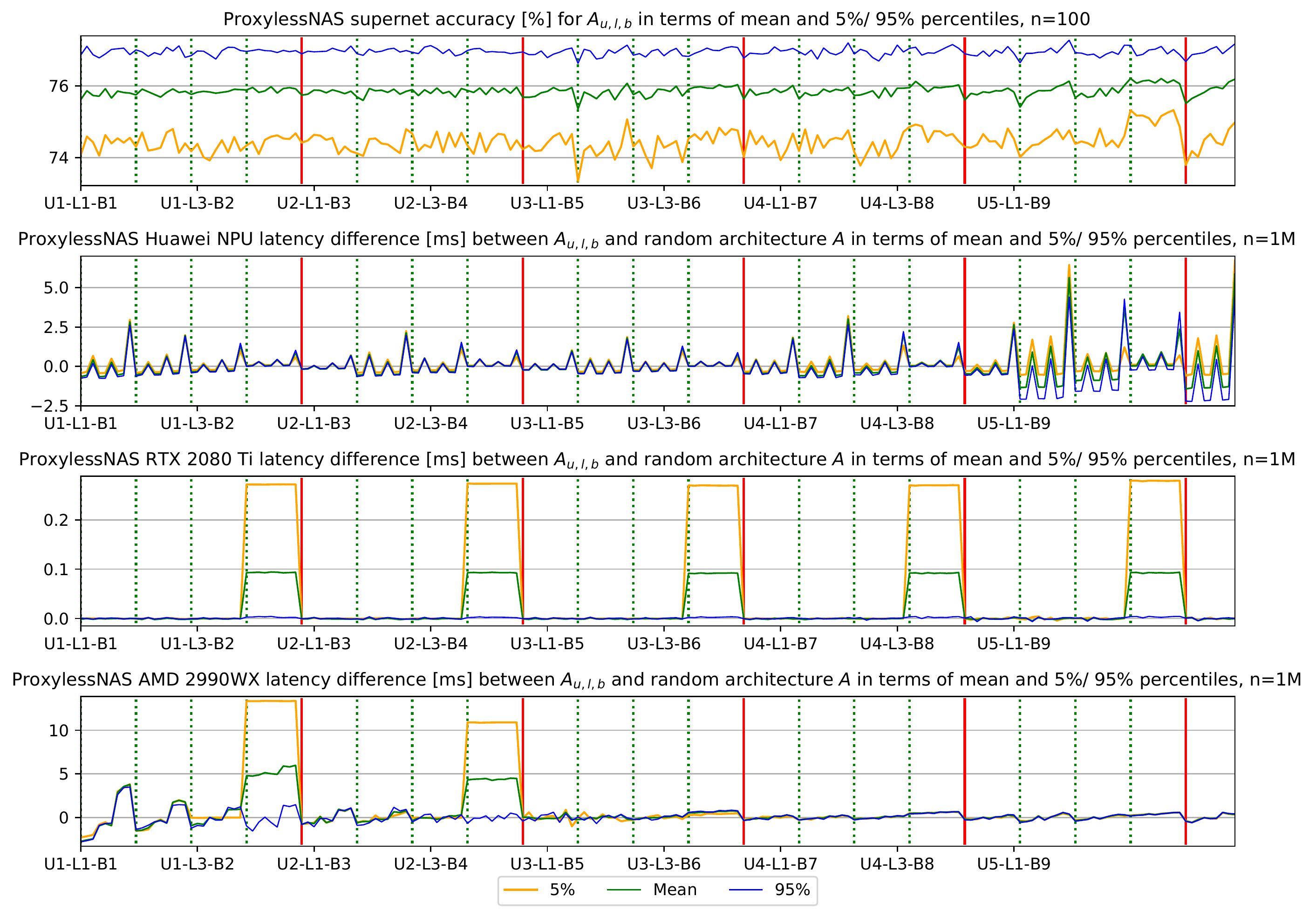}
    \vspace{-8mm}
	\caption{$M_{u,l,b}$ and $M_{\tau, (u,l,b)}$ for ProxylessNAS, where  Unit-Layer-Blk (U-L-B) triplets are arranged in increasing order on x-axis, with solid vertical lines demarcating units and dashed vertical lines demarcating layers.  
	For each metric, we present 3 curves, the mean as well as 5\% and 95\% percentile relative performance according to~\eqref{eq:expectation} and \eqref{eq:quantile}. Input resolution is 224.}
	\vspace{-3mm}
	\label{fig:pn_sweep}
\end{figure*}

\begin{figure*}[t]
    \includegraphics[width=7in]{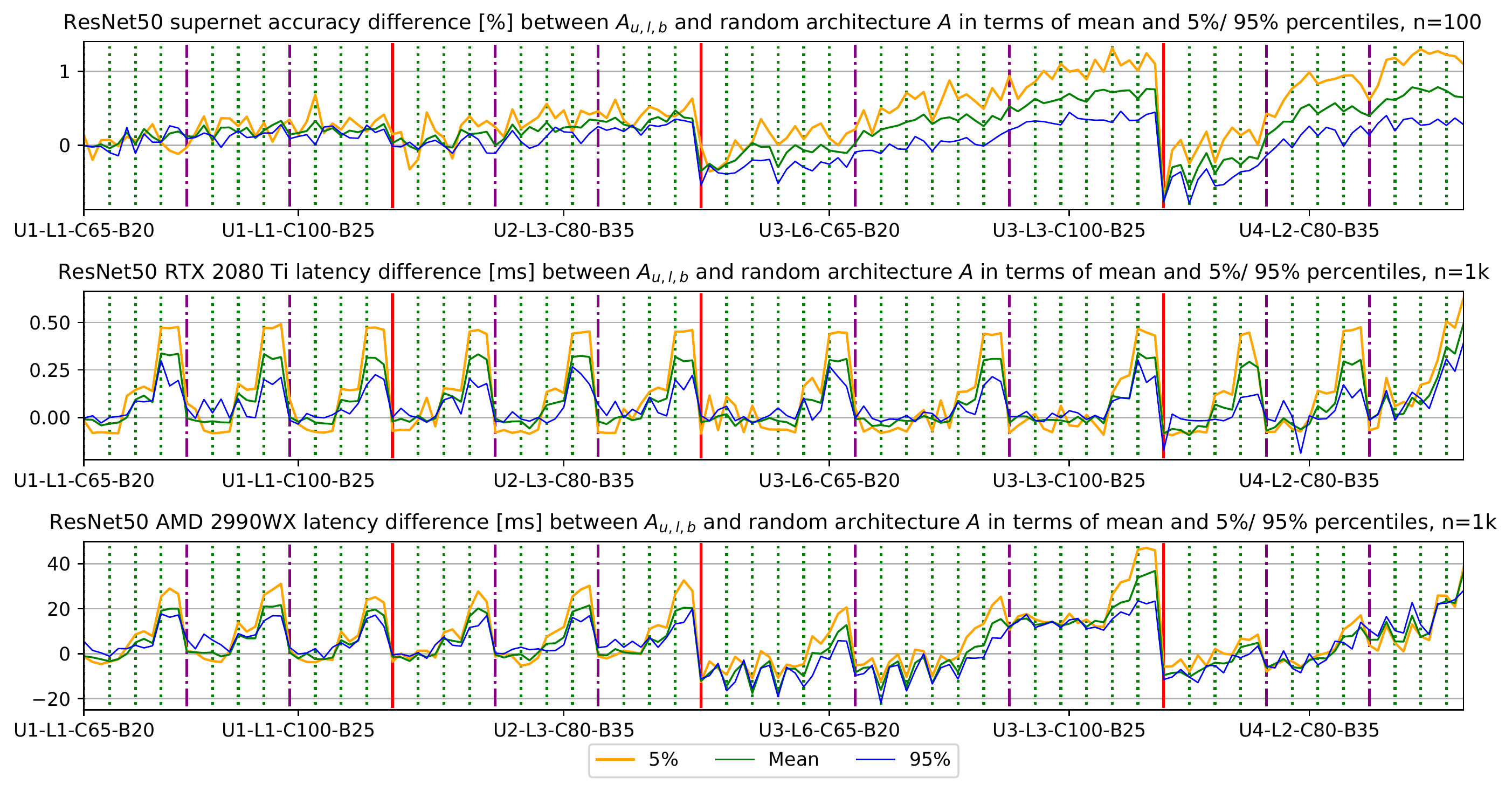}
    \vspace{-8mm}
	\caption{
	$M_{u,l,b}$ and $M_{\tau, (u,l,b)}$ for ResNet50, where  Unit-Layer-Blk (U-L-B) triplets are arranged in increasing order on x-axis, with solid vertical lines demarcating units, dashed lines demarcate unit channel settings, and dotted lines demarcate layers. 
	For each metric, we present 3 curves, the mean as well as 5\% and 95\% percentile relative performance according to~\eqref{eq:expectation} and \eqref{eq:quantile}. Input resolution is 224.}
	\vspace{-3mm}
	\label{fig:rn_sweep}
\end{figure*}

Next, Figure~\ref{fig:pn_sweep} provides results for ProxylessNAS. For accuracy we provide the raw values\footnote{No subtraction term in ~\eqref{eq:expectation} and ~\eqref{eq:quantile}.} instead of the relative deviation as we found it easier to read in the absence of a predictor. While the accuracy curves are less salient than OFA, the mean and 5\%-percentile $M_{\tau,(u,l,b)}$ curves still illustrate a large variation in the latter half of the network. Like OFA, NPU latency varies greatly in the final units, although the sensitivity of unit 4 to block choice is greatly diminished, and is only slightly greater than unit 1. Following OFA, GPU latency is not sensitive to block choice for ProxylessNAS, while CPU latency only varies with expansion ratio in the first unit, likely due to the size of input tensors. 

Finally, Figure~\ref{fig:rn_sweep} illustrates metric profiles for ResNet50. Accuracy is relatively invariant to block choice in the first 2 units but there is a clear, linear dependency on both the number of channels in a unit and the layer expansion ratio in the final 2 units. GPU latency is largely unaffected by block choice while CPU latency greatly depends on channels in units 3 and 4.

\textbf{Sensitivity to network depth.} Figure~\ref{fig:ofa_sweep} shows that when the final layer of a unit (layer 4) is present, accuracy rises uniformly regardless of block choice. This is because including layer 4 ensures that the unit is at maximum length. Moreover, the increase is highest for unit 4, and lowest for unit 1. When designing a search space reduction for OFA, this finding motivates us to ensure that later units remain at maximum length, while earlier units others might be restricted to limit latency. By contrast we do not note a significant rise in NPU latency in the optional layers of any unit. GPU and CPU latency for OFA are largely dependent on unit depth as the presence of a fourth layer causes latency to rise and plateau.

Depth sensitivity for ProxylessNAS is largely the same as OFA, as Figure~\ref{fig:pn_sweep} shows. On the GPU there minimal variation in the amount of added latency for layer 4 of a given unit. To purely minimize GPU latency for MobileNets, all 5 units should be constrained to have 3 layers maximum. However, in practice, this choice should be weighed against potential loss of accuracy. Lastly, CPU latency rises in units 1 and 2 when the fourth layer is present, but only for the mean and 5\%-percentile of architectures. We can therefore infer that most of the random architectures that make up the top-95\%-percentile of CPU latency all have 4 layers in units 1 and 2.

Finally, GPU latency for ResNet50 is sensitive to unit depth, while CPU latency depends on channel size beyond the first 2 units.

%% file: src/application.tex
\section{Application to NAS}
\label{sec:application}
In this section, we demonstrate that knowledge obtained from block profiling can be leveraged to effectively enhance hardware-aware neural architecture search (NAS).
Specifically, we strategically reduce a search space according to insights obtained from block profiling results, then search for the optimal accuracy-latency Pareto frontier.
Experimental results suggest that the insights bring improvements to accuracy-latency tradeoff on various hardware devices.
We also show that when using the insights to maximize accuracy only, we find new architectures with accuracy outperforming the originally published best models found on OFA and ProxylessNAS design spaces. 

\subsection{Insights-Driven Pareto Front Search}
\label{sec:pareto_search_spaces}

We first apply our insights to generate Pareto frontiers for OFA and ProxylessNAS where ImageNet top-1 accuracy is maximized while hardware inference latency is minimized. The details of the pruned search spaces on this task are enumerated below:

\begin{itemize}
    
    \item \textbf{OFA-NPU}: To \textit{reduce latency}, remove kernel 7 blocks for all units. To \textit{increase accuracy}, prioritize optimizing block choice in the final 2 units.

    \item \textbf{OFA-GPU}: To \textit{reduce latency}, constrain units 2, 4 and 5 to have at most 3 layers. To \textit{increase accuracy}, remove MBConv3--3, MBConv3--7 and MBConv4--3 from the search space.

    \item \textbf{OFA-CPU}: To \textit{reduce latency}, constrain units 1, 2 and 3 to have at most 3 layers.
    
    \item \textbf{OFA-Note10-R224/192}: To \textit{reduce latency}, constrain unit 1 to have at most 3 layers. To \textit{increase accuracy}, remove MBConv3--7 and MBConv4--7 from the search space.

    \item \textbf{ProxylessNAS-NPU}: Same as OFA-NPU.

    \item \textbf{ProxylessNAS-GPU}: To \textit{reduce latency}, constrain units 1, 2 and 3 to have at most 3 layers. To \textit{increase accuracy}, remove MBConv3--3, MBConv3--7 and MBConv4--3 from search space. 
    
    \item \textbf{ProxylessNAS-CPU}: To \textit{reduce latency}, constrain units 1 and 2 to have at most 3 layers.
\end{itemize}

Our objective is to demonstrate that our insights can improve search by pruning potentially sub-optimal regions in the design space. While any reasonable search algorithm can be chosen to facilitate this goal, we conduct our search using an Evolutionary Algorithm (EA), a typical family of search algorithms widely adopted in existing NAS practice~\cite{elsken2018efficient, dai2020fbnetv3}.

We start the search with a set of randomly sampled architectures as the initial population.
Next, unit-wise random mutations are performed on architectures from this population, where the choices include adding or removing a layer to/from a stage, as well as changing the block type in an existing layer. 
After getting a set of new architectures, we query the relevant performance metrics, i.e., accuracy and/or hardware latency.
Using the ranks produced by these metrics as the population fitness score, we get the top-performing architectures and merge them with the initial population to create a new top population.
A new round of evolution continues on this new generation. 
When conducting Pareto frontier searches, we set the number of generations to 10, the starting population size to 100 and generate 200 child architectures in each generation. Experimental runs take roughly 1 GPU day per Pareto frontier.

\begin{figure*}[t]
	\centering
	\includegraphics[width=7in]{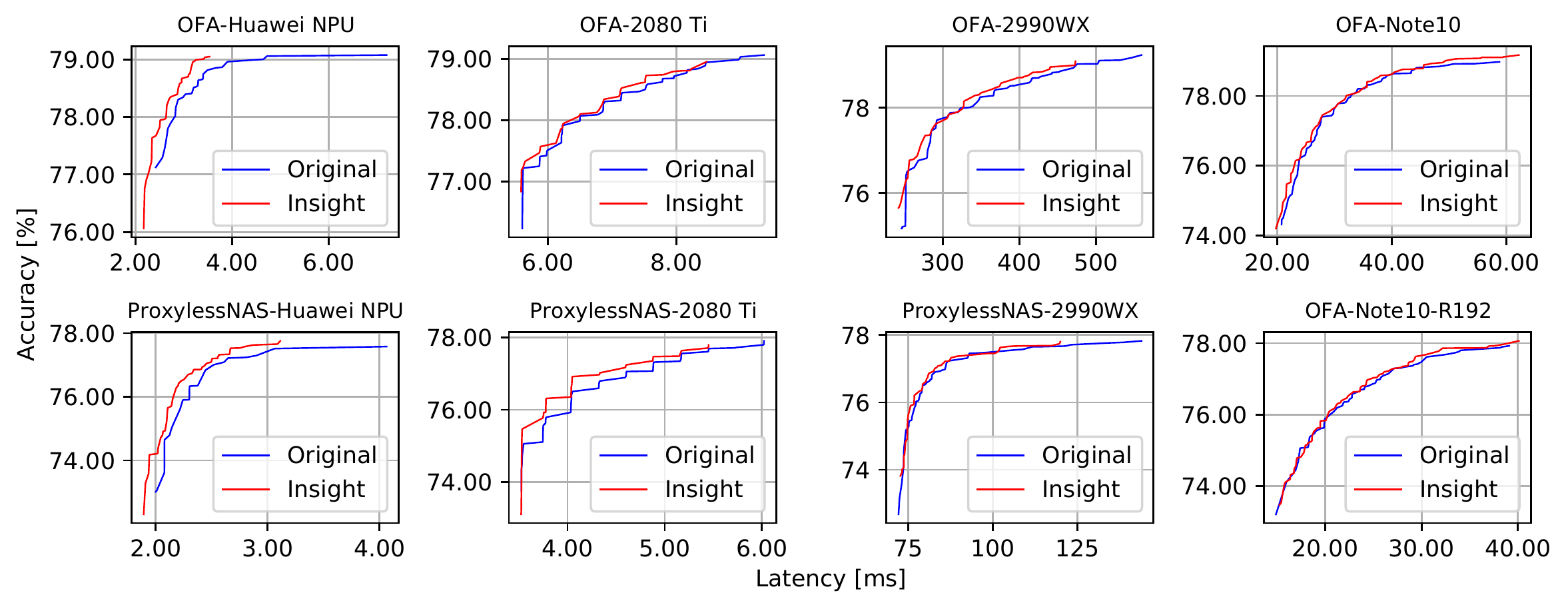}
	\vspace{-7mm}
	\caption{Pareto frontiers contrasting the original search spaces (blue) with our insight-based search spaces (red).}
	\vspace{-3mm}
	\label{fig:Pareto}
\end{figure*}

Figure~\ref{fig:Pareto} shows Pareto frontiers on various hardware platforms.
Under our insight-based search spaces for OFA and ProxylessNAS, the same search algorithm discovers better Pareto frontiers in the mid/low latency regions on NPU, GPU and CPU hardware. 
Noticeably on Huawei NPU, our insights produce up to 3 times latency reduction, while the accuracy is significantly improved for the $<3ms$ region. In fact, to make the insight Pareto curves visible, we removed the highest latency entries for the original spaces on the NPU as they were in excess of 10ms and provided minimal accuracy gain.
For 2080Ti GPU and 2990WX CPU, we use our insights to specifically improve the accuracy of mid/low latency regions, which explains why the Pareto frontiers are shorter on the right end but better in the remaining regions when compared to the full space baseline.

While improvements on the Note10 platform are sparse, our pruned search space achieves higher gains in more places than the original search space. Optimizing for accuracy against Note10 latency is difficult because of how patently high correlated both metrics are. Removing a high latency operation means removing a high accuracy operation at the same time. This is in contrast to other metrics like NPU latency, where one can exploit the unfriendliness of kernel size 7 blocks to reduce latency, or GPU latency, where we can prune low-accuracy operations without an increase in latency due to near constant inference time.

\subsection{Unconstrained Maximum Accuracy Search}
\label{sec:maxacc}
In addition to our Pareto frontier search task, we also leverage our insights to prune the search spaces such that the search algorithm finds high-accuracy architectures more readily. In this situation, we focus on finding the maximum possible accuracy and disregard latency constraints. For the OFA and ProxylessNAS design spaces, we remove all expansion 3 blocks, as well as MBConv4--3 for all units since they are sub-optimal in terms of accuracy. For the ResNet50 design space, we constrain the last two units to always have the maximum number of layers and output channels.

As our goal in this situation is to find a \textit{single} high-performance architecture, rather than a spectrum, we set the number of generations to 4, the starting population size to 20 and generate 50 child architectures in each generation. We do not perform fine-tuning after finding an architecture. Experimental runs take 2 GPU hours for OFA and ProxylessNAS, and roughly 3 GPU hours for ResNet50.

\begin{table}[t]
	\centering
	\caption{Maximum top-1 ImageNet  accuracy search results on different design spaces, compared to existing works. We show averages over 5 random seeds for our experiments.}
	\label{table:max_acc_search_results}
	\begin{tabular}{lcc} 
	\toprule
	    \textbf{Model}                               &\textbf{Accuracy} & \textbf{MACs}  \\ \midrule
	    MobileNetV2 \cite{sandler2018mobilenetv2}    & 72.0             & 300M \\
	    MobileNetV3-Large \cite{howard2019searching} & 75.2             & 219M \\
	    OFA \cite{cai2020once}                       & 76.0             & 230M \\ 
	    OFA$_{Large}$                                & 79.0             & 595M \\ \midrule
	    OFA-insight                                  & \textbf{79.2} $\pm$ 0.04 & 342M  \\
	    OFA-base                                     & 78.9 $\pm$ 0.07 & 292M  \\
	    ProxylessNAS-insight                         & \textbf{77.9} $\pm$ 0.04 & 417M  \\
	    ProxylessNAS-base                            & 77.6 $\pm$ 0.08 & 359M  \\
	    ResNet50-insight                             & \textbf{80.0} $\pm$ 0.03 & 2.81B  \\
	    ResNet50-base                                & 79.9 $\pm$ 0.09 & 2.64B  \\
        \bottomrule
	\end{tabular}
	\vspace{-5mm}
\end{table}

Table~\ref{table:max_acc_search_results} records the results of our maximum accuracy experiments.
Observe that the insight-based search setups consistently find superior networks, both on average and when the maximum accuracy is considered, across different design spaces despite the restrictions imposed. The results imply that the derived insights are more helpful at discovering hardware-friendly architectures for mobile devices, while improving the predictive power of the search spaces. This is most evident for OFA. While the base search space does not exceed the top-1 accuracy of OFA$_{Large}$, the insight-driven search exceeds it by 0.2\%, a significant amount.
Also, not that while the ResNet50 design space achievees the highest accuracy, the number of Multiply-Accumulate operations (MACs) is almost a magnitude higher than OFA and ProxylessNAS spaces. Therefore, we would still recommend the OFA and ProxylessNAS design spaces for mobile applications, e.g. the Samsung Note10 or Huawei NPU, where power or memory consumption is often a concern.

%% file: src/related.tex
\section{Related Work} 
\label{sec:related}

Our work on block and search space profiling draws inspiration from the recent benchmarking efforts in NAS. Benchmarks like NAS-Bench-101~\cite{ying2019nasbench101} and 201~\cite{dong2020nasbench201} define their own search spaces and label the truth accuracy for all possible architectures. They are the first hint that the search space has a significant impact on the performance of an neural architecture. However, the search spaces of NAS-Bench-101 and 201 are relatively small, consisting of 423k and 15k architectures, respectively. In contrast, the OFA-MobileNetV3 search space contains $10^{19}$ architectures~\cite{cai2020once}. Thus, it is too costly to exhaustively evaluate every architecture in such search spaces. A key advantage of our method is that we do not evaluate all possible architectures. Instead, our method enables the profiling of blocks and their placement in units and layers.

There are other works that perform measurements on design spaces. For example, \cite{radosavovic2019network} uses complexity measures to normalize and compare the accuracy error distributions of competing search spaces across parameters or FLOPS. \cite{radosavovic2020designing} uses statistical insights to prune the number of degrees of freedom present in a design space. In contrast, our work uses sampling to obtain block-level insights across different locations in the network, with an emphasis on hardware friendliness. 

More NAS algorithms start to incorporate search space modifications to improve performance. \cite{zhang2020fast} uses different candidate operator sets for each searchable layer, while TinyNAS~\cite{lin2020mcunet} first optimizes the search space to fit hardware resource constraints and then searches for the best architecture.
HourNAS~\cite{yang2020hournas} divides its search space into vital and non-vital components.
P-DARTS~\cite{chen2019progressive} reduces its search space in order to search for large models.
However, these works often lack a systematic analysis about why certain operators are favorable in certain network layers, and in this paper we aim to fill this gap.

%% file: src/conclusion.tex
\section{Conclusion}
\label{sec:conclusion}

In this paper, we propose a sampling methodology to profile neural blocks used in the state-of-the-art neural architecture search spaces of Once-for-All and ProxylessNAS as well as the classical framework of ResNet50.
We profile the blocks in terms of their contributions to end-to-end network accuracy, latency and other metrics across a range of hardware devices, including Huawei Kirin 9000 NPU, Nvidia RTX 2080 Ti GPU, AMD 2990WX CPU and Samsung Note10.  Our findings show that the latency response of each device is unique and helps us to obtain knowledge regarding their sensitivity to block choice, block placement at different depths and macro net structure. By applying our insights to reduce the original search spaces, we find better Pareto frontiers in terms of accuracy and latency, compared to the original search spaces. Finally, we show that the use of insights can also lead to architectures that notably surpass the original best architecture found in OFA on ImageNet top-1 accuracy.